# Improved Sampling for Diagnostic Reasoning in Bayesian Networks

**Mark Hulme**
Computer Science Dept.
RMIT, GPO Box 2476V
Melbourne, Australia, 3001
mrh@cs.rmit.edu.au

## Abstract

Bayesian networks offer great potential for use in automating large scale diagnostic reasoning tasks. Gibbs sampling is the main technique used to perform diagnostic reasoning in large richly interconnected Bayesian networks. Unfortunately Gibbs sampling can take an excessive time to generate a representative sample. In this paper we describe and test a number of heuristic strategies for improving sampling in noisy-or Bayesian networks. The strategies include Monte Carlo Markov chain sampling techniques other than Gibbs sampling. Emphasis is put on strategies that can be implemented in distributed systems.

## 1 Introduction

This paper describe and tests heuristic strategies for improving sampling when performing diagnostic reasoning in large richly interconnected noisy-or Bayesian networks. Emphasis is put on strategies that can be implemented naturally in a distributed system.

Diagnostic inferencing involves reasoning from a set of observations to a set of possible models, hypothesis or causes. Two common examples of diagnostic problems are medical diagnosis and mechanical fault diagnosis. The class of problems involving some component of diagnostic inferencing is not restricted to those labeled as diagnosis. For example, sensory perception (eg vision) and "induction" both involve reasoning from a set of observations to a set of possible models or causes.

Bayesian networks offer great potential for use in automating large scale diagnostic reasoning tasks. Unfortunately reasoning in large richly interconnected Bayesian networks can be difficult. Exact evaluation of a Bayesian network is NP-hard [Cooper, 1990]. Evaluation of a Bayesian network within probably approximately correct bounds is also NP-hard [Dagum and Luby, 1993]. Diagnostic reasoning in Bayesian network causes particular problems.

Methods for evaluating Bayesian networks are typically broken up into two main classes; exact evaluation algorithms and simulation based methods. The exact evaluation algorithms include the Kim-Pearl poly-tree algorithm [Pearl, 1988] and Speigelhalter's clique tree algorithm [Lauritzen and Spiegelhalter, 1988]. All exact evaluation algorithms have exponential time complexity in the number of loops across the width of the network.

There are two main classes of simulation algorithms used in the Bayesian network community; forward sampling and Gibbs sampling. Forward sampling schemes [Henrion, 1986] have good performance even when applied to highly looped networks but only if all the evidence nodes are at the roots of the network. If the network is structured to represent a chain of causal events, as we often want to do when representing diagnostic problems, then forward sampling is efficient at reasoning from the causes to effects. Diagnosis involves reasoning from effect to cause. Forward sampling schemes have been described as having exponential time complexity in the "amount of diagnostic evidence".

Gibbs sampling [Pearl, 1987; Hrycej, 1990; York, 1992] is the main method used when performing diagnostic inferencing in large richly interconnected Bayesian networks. Traditional Gibbs sampling schemes sample a single node at a time. A simple Gibbs sampling scheme would involve choosing a node at random and assigning the node a value according to it's conditional probability given the state of the other nodes in the network. This Gibbs sampling scheme defines a Markov chain where the states of the Markov chain represent states of the free nodes of the network. As long as the Markov chain is aperiodic and contains at most one irreducible closed subset of persistent states, the Gibbs sampler is guaranteed to generate a sample that is representative of the posterior probability of the states of the network in the long run. However, the samples are drawn dependently and may be highly biased in the short run.

In diagnostic problems the causes tend to have small prior probabilities. The effect of this is to set up com-petition between possible causes not to have to explain positive pieces of evidence. A small diagnostic Bayesian network is shown in figure 1a. An earthquake and a boy are considered the only realistic explanations for a vase being broken. Given we know that the vase is broken, one of the two events must have occurred. The small prior probabilities of these two explanations means that it is unlikely that both occurred.

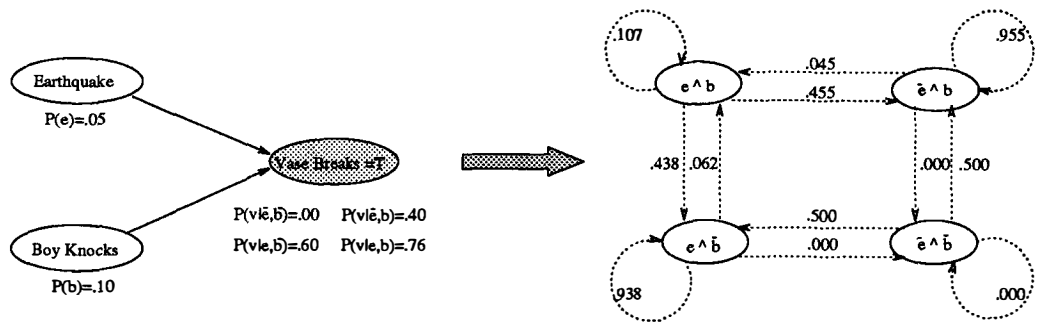

Figure 1: a. Diagnostic Network and b. Gibbs Sampling Markov Chain

The Markov chain defined if we choose one of the free nodes with equal probability and assign it a new value according to its current conditional probability is given in figure 1b. If this Markov chain is traced then it is easy to see that the Gibbs sampler is likely to spend a long time in one of the two high probability states, $e \wedge \bar{b}$ or $\bar{b} \wedge e$, before flipping to the other state. The only path between the high probability states is through the low probability state $e \wedge b$. In many diagnostic reasoning tasks the Markov chain defined by simple Gibbs strategies involve high probability states separated by low probability states. The samples generated by such a Gibbs sampler will be highly dependent and biased in the short run.

The rest of the paper is broken up into four sections. Section 2 describes the diagnostic model examined in this paper and introduces some notation. Section 3 describes the improved sampling strategies investigated in this paper. Subsection 3.1 presents a simple scheme for identifying nodes in the network that need not be simulated. In section 3.2 it is argued that when sampling a node we should only condition on the subset of Markov blanket nodes that d-separate the node from all evidence. Section 3.3 describes the Monte Carlo Markov chain (MCMC) sampling framework. Gibbs sampling is a kind of MCMC sampling [York, 1992]. The framework is used to engineer Markov chains that move between high probability states more easerly. Section 4 presents some empirical results on the strategies described. Section 5 gives some conclusions and suggests some future work.

## 2 Diagnostic Problem

Figure 2 shows the form of the diagnostic problems examined in this paper. The nodes of the graph are split into two sets; model nodes and sensory nodes. The model nodes describe our "world model". We assume that the links between model nodes are structured to represent our views on uncertain causal interaction in the real world. The sensory nodes represent the set of possible direct observations of the world. Links between model nodes and sensory nodes describe how a state of the world effects what we expect to observe.

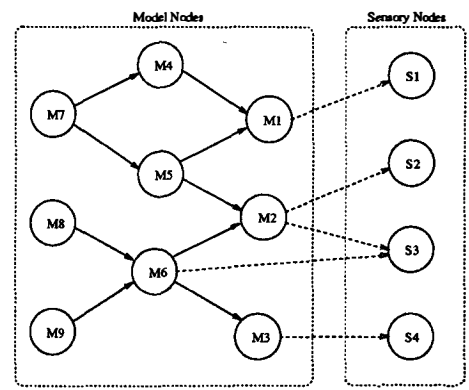

Figure 2: Diagnostic network problem

Nodes of arbitrary type are denoted by the symbol $n$. Model node are denoted using the symbol $m$. Joint states of the model nodes are denoted by the symbol $\theta$. Sensory nodes are denoted by the symbol $s$. The set of observed evidence nodes is referred to with the symbol $Y$.

All nodes in the network are modeled as binary noisy-or units [Pearl, 1988; Peng and Reggia, 1990]. The noisy-or unit is derived when we assume that links represent independent causal events. If we assume that the probability of $n_i$ causing $n_j$ independent of all other events is $p_{ij}$, then the equation below gives the probability that $n_j$ is true given the state of all its direct causes $\pi(n_j)$.

$$P(n_j|\pi(n_j)) = 1 - \prod_{n_i \in \pi(n_j)} (1 - p_{ij})^{n_i}$$

To evaluate this equation we need the set of all possible direct causes of a node. This is not as difficult as it may seem. If the set of direct causes of a node is not sufficiently complete, then we can add an extra node clamped true to represent all other true causes. The network structures drawn do not always explicitly represent these external causes but it is sometimes useful to acknowledge they exist.

Noisy-or units are not the only unit used in Bayesian networks structured to represent chains of causal events. Other examples are noisy-and and noisy-inhibitory units. Noisy-and units model situations where two events need to co-occur to have an uncertain causal effect on another event (eg financial security and friendship lead to happiness). Noisy-inhibitory units model situations where uncertain cause effect relationship may be stopped by a third event (eg contact with flu and flu injection lead to no flu). Noise-or units are the most popular and we will concentrate on networks of these units.

## 3 Improved Sampling

In the next three sections some heuristic strategies are described for improving sampling when performing diagnostic reasoning. The first subsection describes a simple strategy for clamping nodes of the network before simulation begins. In the second subsection we argue that when sampling a network you need only conditioning on the nodes in the Markov blanket that d-separate the node from the evidence. The last section looks at improving sampling of competing nodes by changing the structure of the Markov chain.

### 3.1 Clamping Nodes

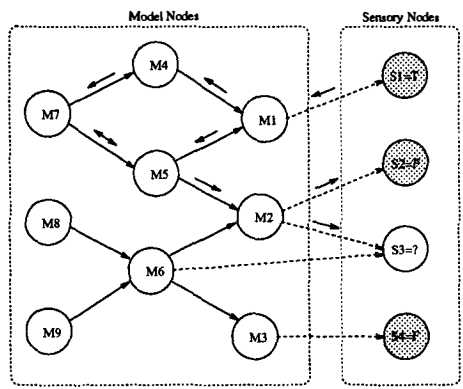

Figure 3: Flow of Positive Evidence

As described in the introduction, the fact that the prior probabilities of model nodes tend to be very small creates problems when Gibbs sampling. This fact also creates opportunities. We can use the fact to help identify nodes of the network that are almost certainly false. If we assume that the prior probability of model nodes approaches zero, then it is easy to see that all nodes that are not ancestors of a true sensory node or descendants of ancestors of a true sensory node are almost certainly false.

It is easy to implement a strategy for identifying which nodes can be clamped in a distributed framework.

1. All nodes are clamped to false.

2. A signal is propagated backwards from all positive sensory nodes to ancestors to indicate that the network has positive diagnostic evidence and they should be unclamped. In figure 3 nodes $m_1$, $m_4$, $m_5$ and $m_7$ receive this signal from $s_1$.

3. A signal is then propagated forward to all descendents nodes to indicate possible increased positive causal support and that they should be unclamped. In figure 3 nodes $m_2$, $s_2$ and $s_3$ will receive a signal from $m_5$ and $m_7$.

No other nodes in a network of noisy-or units receive positive sensory evidential support.

It is important to look at the impact of clamping nodes that are not absolutely false to false. The true posterior probability of the nodes clamped false must be between zero and their prior probability before any evidence arrived. The prior probability of nodes is easy to find by forward sampling. The effect of clamping on unclamped nodes is more difficult to determine. For example, clamping node $m_6$ will have an impact on node $m_5$. This issue is investigated empirically in section 5.

The strategy presented above is easy to extend to networks involving noisy-and units. The forward propagation rule needs be changed so that signals can only be sent forward through a noisy-and unit when all its' parents have received signals that they have positive causal or evidential support. Extension of the rule to inhibitory units would require a more involved signal process. When dealing with inhibitory units negative sensory evidence can increase the probability of model nodes. The fact that someone does not have a flu increases your belief that they have had a flu injection slightly. If we know that there is a reasonable chance that they have had direct contact with someone with a flu then this raises the probability much further.

### 3.2 Sampling the Evidence

As mentioned in the introduction there are two main types of simulation procedures used in Bayesian networks; forward sampling and Gibbs sampling. In forward sampling when a node is simulated only the state of the parent nodes is conditioned on. In Gibbs sampling when a node is simulated all nodes in the Markov

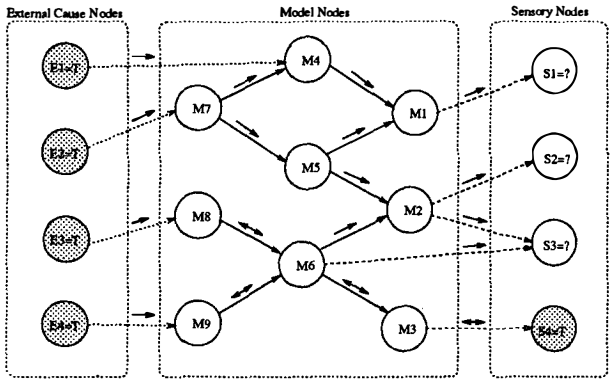

Figure 4: Flow of all Evidence

$$P(n_{ij}|\xi) = \frac{P(n_{ij}|\pi(n_i))\prod_k P(\lambda_k^e(n_i)|\pi_{-n_i}(\lambda_k^e(n_i)), n_{ij})}{\sum_l P(n_{il}|\pi(n_i))\prod_k P(\lambda_k^e(n_i)|\pi_{-n_i}(\lambda_k^e(n_i)), n_{il})}$$

In the equation $\pi(n_i)$ stands for the set of parent nodes of $n_i$, $\lambda(n_i)$ stands for the set of child nodes of $n_i$ and $\lambda^e(n_i)$ is the set of child nodes carrying diagnostic evidence. In figure 3 $m_3$ is the only child node of $m_6$ that carries diagnostic evidence. The states of nodes $s_3$, $m_2$ and $m_5$ should not be considered when simulating this node.

Some caution must be used when applying the equation above. In Gibbs sampling we can visit nodes in any order. In forward sampling nodes are typically visited in a forward order and the state of the network only registered when a complete pass through the network is made. If we are only interested in the marginal probability of nodes then we can visit nodes in any order subject to the restriction that when a node is sampled its' parents have been drawn based on the same causal evidence.

More important than the ability to write down a single equation for sampling nodes is the idea that there is a simple distributed mechanism that can be used to determine how a node should be sampled. As seen in figure 4 all that needs to be done is for the evidence nodes to send signals both forwards and backwards through the network.

### 3.3 Markov Chain Engineering

While the strategies in the previous two subsections improve sampling performance, they do not attack the key problem encountered when trying to perform diagnostic reasoning in noisy-or Bayesian networks. Nodes tend to compete not to have to explain positive evidence. The result of this is that moving between high probability states of the network often requires changing the value of more than one node at a time. This can give diagnostic problem solving a combinatorial flavor.

One strategy designed to help overcome this problem is to merge highly correlated variables into a single variable. This has been suggested in both the statistical literature [Hastings, 1970] and the Bayesian network literature [Pearl, 1988; Jensen et al., 1995]. The strategy is known as blocking. In this subsection we describe the MCMC [Hastings, 1970] framework. It is shown that there is more freedom change the nature of the Markov chain than permanent blocking.

#### 3.3.1 The MCMC Framework

In the MCMC framework [Hastings, 1970] the goal is to set up a Markov chain that when simulated produces a sample representative of the distribution of blanket are conditioned on. Forward sampling will only produce a representative sample if all the evidence nodes are root nodes. The Gibbs sampler will produce a representative sample irrespective of where the evidence nodes are as long as the Markov chain defined by the sampler is irreducible. Intuitively the reason that the Gibbs sampler works irrespective of which nodes are evidence nodes is because it accounts for the evidence coming from all possible directions. This is often inefficient. We might do better to consider the flow of evidence in the net when sampling.

In figure 4 $s_4$ is an evidence node. Learning the value of $s_4$ effects the probability of all ancestor nodes $m_3$, $m_6$, $m_8$ and $m_9$. The effect of the changed probabilities of ancestor nodes also flows forward to all descendants $m_2$, $s_2$ and $s_3$. Node $s_4$ is not the only evidence in the network. Root nodes can be thought of as having true evidence nodes representing true external causes of the node pointing into them. Evidence flows up all branches of the network from behind the root node. Some non-root nodes such as $m_4$ will also be directly attached to a true external cause node.

The arrows on figure 4 show direction of all evidence flows. The nodes $m_3$, $m_6$, $m_8$ and $m_9$ receive evidence from all ancestors and some descendents nodes. This section of the graph should be Gibbs sampled. All other nodes only receive evidence from ancestor nodes and should be forward sampled. Simulation involves visiting each node in turn and calculating probabilities of the values of the node conditioned on the subset of nodes in the Markov blanket that d-separate it from all evidence nodes. The node is assigned a value according to these conditional probabilities.

Recognizing the connection between forward sampling and Gibbs sampling allows us to use one equation to calculating the appropriate conditional probabilities. The probability of node $n_i$ having value $j$ given the subset of Markov blanket nodes, $\xi$, that separate the node from the evidence nodes is given by the equation below.[1]

---

[1] It should be noted that the whole of the Markov blanket of a node needs to be considered if the network is to be annealed.

interest. It is assumed that the goal is to produce a sample representative of the distribution $P(\theta|Y)$. A Markov chain with transition probabilities $P(\theta_i \Rightarrow \theta_j)$ is guaranteed to produce a sample representative of $P(\theta|Y)$ if it contains at most one irreducible closed subset of persistent states, it is aperiodic and it has $P(\theta|Y)$ as its' stationary distribution. The Markov chain has the stationary distribution $P(\theta|Y)$ if it satisfies the equation below.

$$\sum_i P(\theta_i|Y)P(\theta_i \Rightarrow \theta_j) = P(\theta_j|Y) \qquad (1)$$

A sufficient condition for a Markov chain to have $P(\theta|Y)$ as a stationary distribution is that it satisfies the reversibility condition.

$$P(\theta_i|Y)P(\theta_i \Rightarrow \theta_j) = P(\theta_j|Y)P(\theta_j \Rightarrow \theta_i). \qquad (2)$$

Hastings showed that a Markov chain that produces the correct sample when simulated can be constructed from a sequence of Markov chains. Consider a sequence of transition matrices, $(P_1 \ldots P_d)$, and their matrix multiple, $P = \prod_{i=1}^d P_i$. Assume each matrix $P_i$ has $P(\theta|Y)$ as one of stationary distributions. If $P$ defines a Markov chain that is aperiodic and has at most one irreducible closed subset of persistent states, then the Markov chain will produce a sample when simulated that is representative of $P(\theta|Y)$ in the long run. Repeatedly simulating the sequence of transition matrices $(P_1 \ldots P_d)$ will also produce a sample representative of $P(\theta|Y)$ in the long run.

### 3.3.2 MCMC strategies

From the description above it is easy to see that there is considerable freedom in designing MCMC sampling strategies. In order to develop new sampling strategy we need to specify three components.

- First, we need to choose the set of Markov chain structures to be used.
- Second, we need to specify the transition probabilities for the Markov chains.
- Last, we need to define the sequence that the Markov chains are to be simulated in.

We will discuss how each of these components can be chosen to help improve diagnostic sampling below.

**Chain Structures:** If the Markov chains that we develop are to satisfy the reversibility condition, then any chain that has a transition link from $\theta_i$ to $\theta_j$ must also have a reverse link from $\theta_j$ to $\theta_i$. With this constraint we can see that developing a Markov chain involves partitioning the set of all elementary states of our model into fully interconnected subsets.

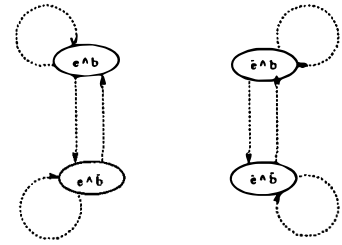

Figure 5: Simulating a Single Variable

Figure 5 shows the Markov chain defined when we sample node $e$ of the broken vase network described in the introduction. As already discussed, if we simulate this network by only allowing the value of one variable to change at a time, then our sampler is likely to generate a highly biased sample in the short term.

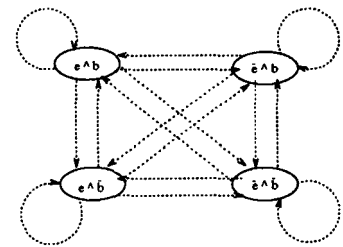

Figure 6: Blocking a Pair

Figure 6 shows the effect of blocking the two competing causes in the broken vase example. The Markov chain now allows for direct movement between the two high probability states $e \wedge \bar{b}$ and $\bar{b} \wedge e$ and will therefore produce a much better quality sample. The problem with blocking is that it is computationally expensive. Simulating a single binary involves inspecting the Markov blanket of the node twice. Simulating a blocked pair of binary nodes involves conditioning on the joint Markov blanket of the node pair four times. The joint Markov blanket of a pair of nodes may be up to twice as large as the Markov blanket of a single node. In the worst case simulating $n$ binary nodes involves conditioning on the joint Markov blanket of the $n$ nodes $2^n$ times.

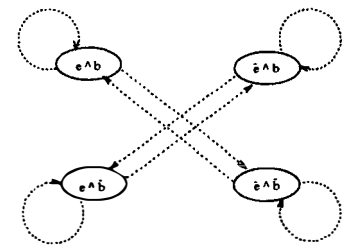

Figure 7: Swapping a Pair

An alternative to blocking pairs of nodes is to swap the value of nodes at one time. Figure 7 shows the Markov chain defined when we allow the two competing causes in the broken vase network to swap their values at the same time. Again there is a direct path between the two high probability states $e \wedge \bar{b}$ and $\bar{b} \wedge e$. The Markov blanket of the two nodes has to be inspected twice for each simulation and the size of the Markov blanket is up to twice as large. This gives an equivalent of up to four single node Markov blanket calculations per simulation. In this paper we restrict ourselves to single node, paired node blocking and paired node swapping simulations.

**Transition Probabilities:** The transition probabilities attached to the Markov chain need to be specified so that the Markov chain satisfies the reversibility condition. Hastings [Hastings, 1970] derived a formula for the range of possible assignments of the transition probabilities. In this paper if the Markov chain allows us to transfer between a set of states $\Theta_m$, then we take the transition probability of moving between $\theta_i \in \Theta_m$ and $\theta_j \in \Theta_m$ as $P(\theta_i|\Theta_m, Y)$. This reduces to the normal Gibbs sampling transition probabilities when sampling one node at a time. It is easy to show that this assignment of transition probabilities satisfies the reversibility condition.

$$\begin{aligned} P(\theta_i|Y)P_m(\theta_i \Rightarrow \theta_j) &= \frac{P(\theta_i|Y)P(\theta_j|Y)}{\sum_{\theta_l \in \Theta_m} P(\theta_l|Y)} \\ &= P(\theta_j|Y)P_m(\theta_j \Rightarrow \theta_i) \end{aligned}$$

The equation below shows how this transition probability can be calculated. The symbol $\Delta_m$ stands for the set of all random variables that do no have the same assignment in all elements of the set of states $\Theta_m$. The symbol $n_k^i$ stands for the statement that the $k$ node has the value given to it in $\theta_i$ state.

$$P(\theta_i \Rightarrow \theta_j) = \frac{\prod_{n_k \in \Delta_m \cup \lambda(\Delta_m)} P(n_k^i|\pi(n_k^i))}{\sum_{\theta_j \in \Theta_m} \prod_{n_k \in \Delta_m \cup \lambda(\Delta_m)} P(n_k^j|\pi(n_k^j))}$$

It is sometimes argued that method of assigning transfer probabilities developed by Metropolis is more efficient than the one presented here because it encourages more transitions of the Markov chain during simulation. If $\Theta_m = \{\theta_i, \theta_j\}$, then the transition probability from $\theta_i$ to $\theta_j$ is given by $\min(1, P(\theta_j|Y)/P(\theta_i|Y))$. The two methods of assigning transition probabilities are compared empirically in the next section.

**Sequence of Markov Chains:** When choosing a sequence of Markov chains to use in simulation we have to ensure that it is alway possible for the resulting sampler to move between all non-zero probability states within a finite number of moves. The choice of sequence of Markov chains should be made in order to minimize the time it takes to sample the network and maximize the quality of the sample produced. The following strategies are designed to ensure that blocking and swapping which is more expensive than ordinary sampling are used only where it is most needed.

- Spouse nodes that receive positive diagnostic evidence from a true sensory node are paired randomly and block sampled. All other nodes are sampled normally.
- Nodes that share a true common child are paired at random and block sampled. All other nodes are sampled normally.
- Spouse nodes that receive positive diagnostic evidence from a true sensory are randomly paired and swap sampled in 80% of the time. Any node that is not swap sampled is sampled normally.
- Nodes that share a true common are paired at random and swap sampled in 80% of the time. Any node that is not swap sampled is sampled normally.

When choosing a sequence of Markov chains it might also be useful to include chains that simulate difficult to simulate areas of the network more often.

## 4 Empirical Results

In this section some empirical results are presented on the different sampling strategies described in the paper. The goal of this study is to see how well the strategies work at producing rough posterior probabilities for model nodes in a limited time. It is assumed that it is better for an algorithm to give a rough indication of the posterior probability of all model nodes, rather than a precise indication of the probabilities of most model nodes. The two goals of robustness and precision can sometimes conflict. An estimate of the posterior of a model node, $m_i$, is considered accurate enough if it is within $\pm\sqrt{P(m_i|Y)(1 - P(m_i|Y))}/5$ of the true probability, $P(m_i|Y)$. Notice that this bound allows probabilities near 0.5 to be estimated with less precision than probabilities near zero or one.

The strategies were tested on an engine fault network diagnosis network modeled using the fault diagnosis charts found in a standard car manual as a base [Gregory's, 1990]. The network has 187 model nodes, 82 sensory nodes and 600 links. All nodes are considered to be linked to bias nodes which represent other causes. Five test cases were used. The number of pieces of evidence in these cases ranged from 4 to 20. The number of pieces of positive evidence ranged from 2 to 9. Each test case was run at least 20 times per strategy.

Posterior probabilities of model nodes were estimated by the average probability of the node being turned on whenever a Markov chain is applied that would allow the node to change value. The same strategy is used in the stochastic simulation with Markov Blanket scoring algorithm introduced in [Pearl, 1987].

|  | Time | 5 Runs | 500 Runs | 1000 Runs | 2000 Runs |
|---|---|---|---|---|---|
| Gibbs sampling | 1.00 | 51.9 | 51.5 | 50.8 | 50.0 |
| Gibbs sampling and clamping | 0.81 | 52.1 | 51.0 | 49.6 | 49.3 |
| Gibbs sampling and forward sampling | 0.48 | 51.7 | 46.4 | 43.6 | 34.3 |
| Block spouses if within cover | 1.30 | 51.9 | 48.6 | 44.5 | 36.2 |
| Block spouses if a parent is true | 1.26 | 52.0 | 43.1 | 39.2 | 35.1 |
| Swap spouses if within cover | 1.16 | 52.0 | 44.1 | 40.1 | 34.8 |
| Swap spouses if a child is true | 1.13 | 52.3 | 47.2 | 44.5 | 36.8 |
| Metropolis sampling | 1.02 | 52.2 | 51.7 | 51.2 | 50.1 |
| Optimized with random selection | 0.57 | 52.2 | 30.7 | 18.3 | 11.3 |
| Optimized with forward and backward pass | 0.38 | 51.8 | 35.4 | 22.0 | 11.0 |

Table 1: Average Error Rates on Test Problem

A summary of the results is given in table 1. It shows the comparative time each strategy took and the average number of errors made after varying numbers of runs. When interpreting these results it is important to remember that the efficiency and effectiveness of each strategy is going to be dependent on the network, the cases, hardware and software used. Testing of the strategies across a broader range of networks and cases is needed in the long term. The results presented in table 1 are discussed below.

### 4.1  Gibbs Sampling

The first row of the table shows the performance of Gibbs sampling on the test problems. Little progress is made towards solving the problems. The error measure shown in the table is a bit deceptive. Gibbs sampling occasionally performs reasonably well on one of the test problems, but performs badly on the great majority. This is an indication that the Gibbs sampler often gets stuck in bad local minima. Similar problems are reported in [Jensen et al., 1995].

### 4.2  Clamp Nodes

The second row of the table shows the average number of errors after selected model nodes have been clamped according to the strategy described in section 3. The clamping strategy does not appear to lead to any significant increase or decrease in the quality of the sample. The main benefit that comes from node clamping is the reduction in computational time. The node clamping strategy took 19% less time than Gibbs sampling.

### 4.3  Sampling from the Direction of Evidence

The third row of the table shows the effect of recognizing that some nodes do not receive any diagnostic evidence. This is the first strategy to make any progress towards solving the test cases within the 2000 run limit. Not only is quality of the sample improved but the time taken to sample the network is reduced 50% on the test cases.

### 4.4  Blocking and Swapping Nodes

Rows four to seven of the table compare the various strategies for sampling competing nodes. All strategies lead to similar significant improvements in the quality of the sample. There is no detectable difference between the blocking and swapping strategy on the test cases. All blocking and swapping strategies take more time to run than Gibbs sampling. The most time efficient strategy is to swap sibling nodes in pairs when a spouse is true.

### 4.5  Choosing Transfer Probabilities

The eight rows of the table shows Metropolis sampling. This is equivalent to Gibbs sampling except Metropolis transfer probabilities are used (see section 3). The runs shown in table 1 indicate that there is very little difference in performance between the two methods. Runs done on different problems show a slight difference in favor of Metropolis sampling.

### 4.6  Concentrating Sampling

The last two rows of the table show combined strategies. Both strategies involve using Metropolis transfer probabilities in the Markov chain, node clamping, sampling from the direction of the evidence and spouse swap sampling when a parent is on. The first optimized strategy visits nodes in a random order. The second optimized strategy alternates between a forward pass and a backward pass through the network. In the second strategy nodes that do not receive any diagnostic evidence are only sampled on the forward pass. The argument for this is that node that receive diagnostic evidence tend to be much more difficult to simulate and should attract more of the computational resources. The forward-backward pass strategy produced the same quaility samples as the random selection strategy. The strategy runs in 66% of the time of the random selection strategy.

### 4.7 Comment

The combined strategies go a long way towards solving the test cases within the 2000 runs. The combined strategies produce much higher quality samples than straight Gibbs sampling and in less than half the time.

## 5 Conclusion and Further Work

In this paper we demonstrated that by being more sensitive to the structure of the network and the nature of the problem to be solved the efficiency of the sampling procedure can be dramatically improved. All the suggestions in this paper could implemented in a distributed framework.

There is scope for further work in at least two areas. First, more work could be done on determining when it is worth simulating a node. The range of nodes that need to be simulated could be narrowed down further by recognizing that nodes that are not ancestors of *unexplained* true nodes, descendants of *unexplained* true nodes or descendants of ancestors *unexplained* true nodes may be treated as false. Consider the following scenario. [2]

> A car arrives at a gas station. Two possible explanations for this are that the car is in need of gas or repairs. If steam is seen rising from the engine and the driver says that the car is overheating, then the reason for the cars arrival is explained. It is unlikely to be worth considering other explanations for the car arriving at the gas station. If modeled in a network, then any node that is not a possible cause of the observation that the car is overheating, a possible consequence of the car overheating or a consequence of a possible cause of the car overheating may not be worth simulating. After noticing that the radiator is leaking it may be possible to focus simulation efforts further.

The focusing mechanism described above would be useful when modeling a broad domain at varying levels of granularity.

Second, there is plenty of room to develop more creative Markov chain engineering strategies. There is no need to restrict the shape of the Markov chains to single variable, paired blocking and paired swapping chains. There is no need to change the shape of the Markov chain used to simulate a subset of nodes just depending on the state of child or descendant nodes. The shape of chain used to simulate a subset of nodes could be varied according to the state of all surrounding nodes or conditional probabilities on the links of the network. The chain itself may be defined in terms of a heuristic ranking of the states of a subset of nodes. There is considerable work still to be done in exploring the space of different Markov chain engineering strategies.

Some preliminary results on further sampling improvement strategies can be found in [Hulme, 1995].

---

[2] Note that in section 3 we were interested in true sensory nodes. Here we may also be interested in any model nodes that are almost certainly true. Simulation may need to be performed to see if a node is almost certainly true.